# Abstract Learning via Demodulation in a Deep Neural Network

Andrew J.R. Simpson [#1]

[#] *Centre for vision, speech and signal processing (CVSSP), University of Surrey, Guildford, Surrey, UK*
[1] `Andrew.Simpson@surrey.ac.uk`

*Abstract*—**Inspired by the brain, deep neural networks (DNN) are thought to learn abstract representations through their hierarchical architecture. However, at present, how this happens is not well understood. Here, we demonstrate that DNN learn abstract representations by a process of demodulation. We introduce a biased sigmoid activation function and use it to show that DNN learn and perform better when optimized for demodulation. Our findings constitute the first unambiguous evidence that DNN perform abstract learning in practical use. Our findings may also explain abstract learning in the human brain.**

*Index terms*—**Deep learning, abstract representation, neural networks, demodulation.**

## I. INTRODUCTION

Deep neural networks (DNN) are state of the art for many machine learning problems [1], [2], [3], [4], [5], [6]. The architecture of deep neural networks is inspired by the hierarchical structure of the brain [7]. Deep neural networks feature a hierarchical, layer-wise arrangement of nonlinear activation functions (neurons) fed by inputs scaled by linear weights (synapses). Since the brain is adept at abstraction, it is anticipated that deep neural architecture might somehow capture abstract representations. However, it is not presently known how (or even if) abstract learning occurs in a DNN.

One possible way to engineer the process of abstraction is known as *demodulation*. Consider a 1 kHz sinusoidal carrier signal that is multiplied with a 10 Hz sinusoidal modulation signal. This shapes the envelope of the carrier signal into a representation of the modulation signal, and this allows the slower modulation signal to pass through a band-pass circuit that would not otherwise support the low frequency information. In this case, the abstract information (i.e., *about* the 1 kHz carrier) is that the carrier is modulated at 10 Hz. Recovery of the modulation signal is achieved via a nonlinear demodulation operation.

In this paper, we take a sampling theory perspective [8] and we interpret the nonlinear activation function of the DNN as a demodulation device within the context of the well-known MNIST [4] hand-written character classification problem (example image in Fig. 1).

We consider the archetypal sigmoid activation function $[y = 1/(1 + \exp(-x))]$. In order for the sigmoid to act as a demodulator, the input data must be asymmetrical. Taking the example of the sinusoidal carrier signal multiplied with the sinusoidal modulator, if the sinusoids are centred within the sigmoid there will be exactly zero demodulation because the demodulation energy exactly cancels out. However, if the sigmoid is biased, then the energy does not cancel out. Thus, only asymmetrical signals may be demodulated by this method.

The sigmoidal activation function comprises a zero-centred sigmoid which is mapped to the range [0, 1]. As a result, demodulation is not symmetrical. We added a bias term ($\beta$) to the sigmoid to illustrate this;

$$y = \frac{1}{1+\exp(-x-\beta)} \quad (1)$$

We set *x* to be a discrete sampled signal comprising the pair of carrier/modulator multiplied sinusoids as described above, where the modulator frequency was $\omega$. Computing *y* for different values of $\beta$, we used the Fourier transform (of *y*) to compute a modulation energy power spectrum, *H*, with *N* bins. From this power spectrum we computed a ratio of the power of the demodulated signal to the average power in the FFT as follows;

$$g = \frac{H(\omega)}{\frac{1}{N}\sum_{i=1}^{N} H_i} \quad (2)$$

Where *g* is the overall measure of demodulation utility and where $\omega$ is the modulation frequency in question. This ratio is unscaled. However, real networks have noise floors, so we include a subtractive logarithmic term to represent the loss in overall power capable of relating the overall modulation power to the 'overall noise floor' (meaning that an infinitely large ratio at an infinitely small energy is not useful). We extend Eq. 2 as follows;

$$g = \frac{H(\omega)}{\frac{1}{N}\sum_{i=1}^{N} H_i} - \alpha \left(\log 10(\frac{1}{N}\sum_{i=1}^{N} H_i)\right)^{\kappa} \quad (3)$$

where $\alpha$ represents an arbitrary scalar on the subtractive term and $\kappa$ raises the order of the term (i.e., to account for more complex network related effects). We set $\alpha$ to $1.8 \times 10^{-13}$ and $\kappa$ was set to 12 in order to provide some punitive drop-off within the range in question.

This allows us to visualize the point at which the gains in ratio are overcome by the losses in overall gain. Fig. 2a plots the resulting functions of $\beta$ for Eq. 2 (grey line) and Eq. 3 (superposed blue line). At zero bias ($\beta = 0$) there is no demodulation power. At negative bias ($\beta < 0$) there is very little demodulation power but at positive bias the demodulation ratio is asymptotic by around $\beta = 6$ and this is then pulled back by the overall gain term (Eq. 3). Hence, the optimal value of $\beta$ is around 6 (depending on the strength of the subtractive overall gain term).

We applied the same activation function (Eq. 1) to a typical feed-forward DNN, which we used to solve the well-known problem of hand-written character recognition on the MNIST dataset [4]. We evaluated both learning and overall performance of the network as a function of $\beta$. This allowed us to assess whether demodulation was a factor in DNN learning, and hence allowed us to assess the question of abstract learning in a quantitative way.

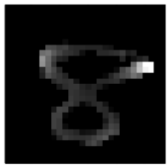

**Fig. 1. Example MNIST image.** We took the 28x28 pixel images and unpacked them into a vector of length 784 to form the input at the first layer of the DNN.

## II. METHOD

We chose the well-known computer vision problem of hand-written character classification using the MNIST dataset [3], [1], [4]. For the input layer we unpacked the images of 28x28 pixels into vectors of length 784. An example digit is given in Fig. 1. Pixel intensities were normalized to zero mean.

Using the biased sigmoid activation function, we built a fully connected network of size 784x784x10 units, with a 10-unit softmax output layer, corresponding to the 10-way digit classification problem. Separate instances of the model were independently trained at various values of $\beta$, using stochastic gradient descent (SGD). However, each individual instance of the model was trained from the exact same random seed. This means that the weights of the network were always initialized to the same (random) values and that the same (randomly chosen) paths were taken during SGD. Hence, changes in performance can be attributed to the different values of $\beta$.

We trained the various instances of the model on the 60,000 training examples from the MNIST dataset [4]. Each iteration of SGD consisted of a complete sweep of the entire training data set. The resulting models were tested on the 10,000 separate test examples at various (full-sweep) iteration points. Models were trained without dropout.

## III. RESULTS

Fig. 2b plots classification error as a function of $\beta$, applied to classifying the separate test data (10,000 examples), after one iteration of training (SGD). The shape of the function matches well the demodulation (Eq. 3) function of Fig. 2a and the two functions are highly correlated ($r = -0.74$, $P < 0.001$, *Spearman*). The classification error function shows the same asymmetrical shape and minima around $\beta = 6$. This demonstrates that demodulation, and hence abstraction, plays a role in the learning and performance of the model.

Fig. 2c plots classification error as a function of iteration number for $\beta = 0$ (representing a traditional sigmoidal activation function) and $\beta = 6$. The biased model ($\beta = 6$) learns more rapidly and performs around 80% better. Both functions appear to have converged, so it would not appear that the difference in results can be attributed to different learning rates. Indeed, it appears that the difference in performance is sufficiently fundamental that extra training would not be able to account for the difference. These results confirm that demodulation, and hence abstraction, plays a key role in learning and performance.

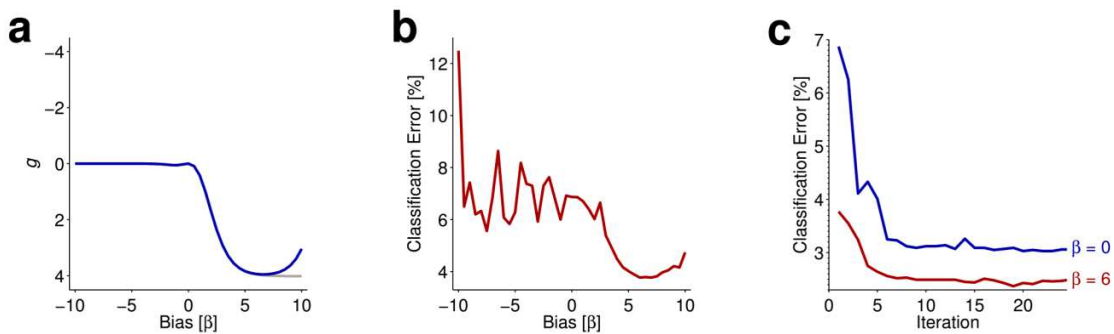

**Fig. 2. Optimal demodulation equals optimal learning. a** Idealized demodulation performance; utility function ($g$ of Eq. 2 and Eq. 3) of sinusoidal modulator demodulated from sinusoidal carrier as a function of bias ($\beta$). The faint grey line represents Eq. 2 and the superposed blue line represents Eq. 3. *Note units are arbitrary and the y-axis is inverted so as to allow comparison with panel b*. **b** Classification error in the test set after the first iteration (an iteration indicates a full sweep of SGD) for different degrees of bias ($\beta$). The two functions are strongly correlated ($r = -0.74$, $P < 0.001$, *Spearman*), indicating that the DNN has performed abstract learning through demodulation. **c** Classification error in the test as a function of training iterations for bias of $\beta = 0$ and $\beta = 6$.

## IV. Discussion

We have set out some simple theory describing how DNN might perform abstract learning through demodulation. We have also introduced a biased sigmoidal activation function and we have characterized the demodulation of this function in both idealized and practical demonstrations. The optimum bias of the practical DNN example matched the theoretical optimum bias for demodulation and provided greatly improved performance in the model. These results provide the first unambiguous evidence that DNN learn abstract representations through demodulation.

In this context, the alternate nonlinear activation functions (abs, tanh, rectified linear unit - ReLU, etc [6]) may be understood as demodulators with different properties. It has been observed that models employing the ReLU function outperform sigmoid activated models [6] in ways that are similar to the results in Fig. 2c. However, it has also been observed that such models require far greater degrees of regularization (such as dropout [6]), which the biased sigmoid activation function does not appear to require. Assuming that demodulation performance is the reason for the improved performance in both cases, the difference in terms of need for regularization is likely due to the high order distortion generated by the abrupt nonlinearity (of the ReLU) and its increased potential for aliasing [8]. Thus, the low-order (smooth) nonlinearity of the biased sigmoid may prove to learn as fast as the ReLU but require less regularization. Future work might characterize the various activation functions in terms of their properties and trade-off of both demodulation performance versus distortion/aliasing, overfitting and regularization requirements.

We have noted that the unbiased sigmoid is, in the idealized case, incapable of performing abstraction by demodulation. Thus, it might be asked: how do these deep networks succeed at all with the unbiased sigmoid? The likely answer to this question is that real data are almost never symmetrical, and hence some small degree of bias would be sufficient to provide abstract learning (e.g., at the level shown in this paper). Indeed, the data employed here is decidedly asymmetrical (even after zero-mean normalization). Thus, it may be generally concluded that traditional sigmoid-activated networks are only capable of abstract learning when the data is asymmetrical. It is also evident that this class of demodulator is inherently sensitive to the polarity of the data, and it may be that some data is more informative in one direction than the other.

More generally, given that the DNN takes its inspiration from the brain, it is interesting to note that the bias scheme described here is not necessary in order for the equivalent brain network to perform abstract learning because the sigmoid activated neurons of the brain are not signed or zero-mean operated [7]. Hence, it would appear that the brain is ideally suited to perform abstract learning via demodulation.

## V. Conclusion

In this paper we have provided evidence that deep neural networks perform abstract learning through a process of demodulation that is sensitive to asymmetry in the data. We have introduced a biased sigmoid activation function that is capable of improved learning and performance. We have shown that the optimum bias point in a practical model matches well that of the idealized demodulation example and that the optimally biased model is fundamentally superior to the traditional sigmoid activated model. These results have broad implications for how deep neural networks are interpreted, designed and understood. Furthermore, our findings may provide insight into the exceptional abstract learning capabilities of the human brain.

## Acknowledgment

AJRS was supported by grant EP/L027119/1 from the UK Engineering and Physical Sciences Research Council (EPSRC).

.